\documentclass[letterpaper, 10 pt, conference]{ieeeconf}  

\IEEEoverridecommandlockouts                              

\usepackage{graphics} 
\usepackage{epsfig} 

\title{\LARGE \bf
Vision and Learning for Deliberative Monocular Cluttered Flight
}
\author{Debadeepta Dey$^{1}$, Kumar Shaurya Shankar$^{2}$, Sam Zeng$^{3}$, Rupesh Mehta$^{4}$, M. Talha Agcayazi$^{5}$, \\ Christopher Eriksen$^{6}$, Shreyansh Daftry$^{7}$, Martial Hebert$^{8}$, J. Andrew Bagnell$^{9}$
\thanks{${1}$, ${2}$, ${3}$, ${7}$, ${8}$ and ${9}$ are with The Robotics Institute, Carnegie Mellon University, Pittsburgh, PA, USA \{debadeep, kumarsha, samlzeng, daftry, hebert, dbagnell\}@ri.cmu.edu}%
\thanks{${4}$ worked on the project while a student at Carnegie Mellon University, currently at NVIDIA Corporation, Santa Clara, CA, USA  rupeshm@nvidia.com,}%
\thanks{${5}$ worked on the project while on an internship at Carnegie Mellon University, George Mason University, Fairfax, VA, USA, magcayaz@gmu.edu}%
\thanks{${6}$ worked on the project while on an internship at Carnegie Mellon University, Harvey Mudd College, Claremont, CA, USA, ceriksen@hmc.edu}%
}

\usepackage{graphics} 
\usepackage{epsfig} 
\usepackage{mathptmx} 
\usepackage{times} 
\usepackage{amsmath} 
\usepackage{amssymb}  
\usepackage{url}
\usepackage{float}

\renewcommand{\baselinestretch}{0.95}

\begin{document}

\maketitle
\thispagestyle{empty}
\pagestyle{empty}

\begin{abstract}
Cameras provide a rich source of information while being passive, cheap and lightweight for small and medium Unmanned Aerial Vehicles (UAVs). In this work we present the first implementation of receding horizon control, which is widely used in ground vehicles, with monocular vision as the only sensing mode for autonomous UAV flight in dense clutter. We make it feasible on UAVs via a number of contributions: novel coupling of perception and control via relevant and diverse, multiple interpretations of the scene around the robot, leveraging recent advances in machine learning to showcase anytime budgeted cost-sensitive feature selection, and fast non-linear regression for monocular depth prediction. We empirically demonstrate the efficacy of our novel pipeline via real world experiments of more than $2$ kms through dense trees with a quadrotor built from off-the-shelf parts. Moreover our pipeline is designed to combine information from other modalities like stereo and lidar as well if available.
\end{abstract}

\section{Introduction}
\label{introduction}
Unmanned Aerial Vehicles (UAVs) have recently received a lot of attention by the robotics community. While autonomous flight with active sensors like lidars has been well studied \cite{scherer2008}, flight using passive sensors like cameras has relatively lagged behind. This is especially important given that small UAVs do not have the payload and power capabilities for carrying such sensors. Additonally, most of the modern research on UAVs has focussed on flying at altitudes with mostly open space \cite{dey2011cascaded}. Flying UAVs close to the ground through dense clutter \cite{ross2013learning} has been less explored. 

\begin{figure}[htbp!]
  \centering
    \includegraphics[width=0.4\textwidth]{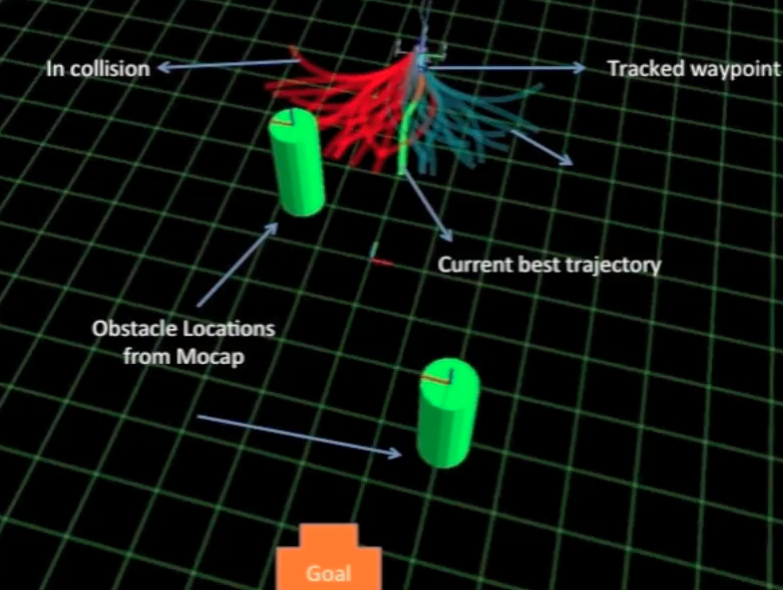}
   \caption{Receding horizon control on UAV in motion capture. A library of $78$ trajecories of length $5$ m are evaluated to find the best collision-free trajectory. This is followed for some time and the process repeated.}
   \label{rec_horz_illus_mocap}
\end{figure}

Receding horizon control \cite{kelly2006perceptor} is a classical deliberative scheme that has achieved much success in autonomous ground vehicles including five out of the six finalists of the DARPA Urban Challenge \cite{urban2008}. Figure \ref{rec_horz_illus_mocap} shows an illustration of receding horizon control on our UAV in motion capture. In receding horizon control, a pre-selected set of dynamically feasible trajectories of fixed length (the horizon), are evaluated on a cost map of the environment around the vehicle and the trajectory that avoids collision while making most progress towards a goal location is chosen. This trajectory is traversed for a bit and the process repeated again. 

\begin{figure}[htbp!]
  \centering
   \includegraphics[width=0.4\textwidth]{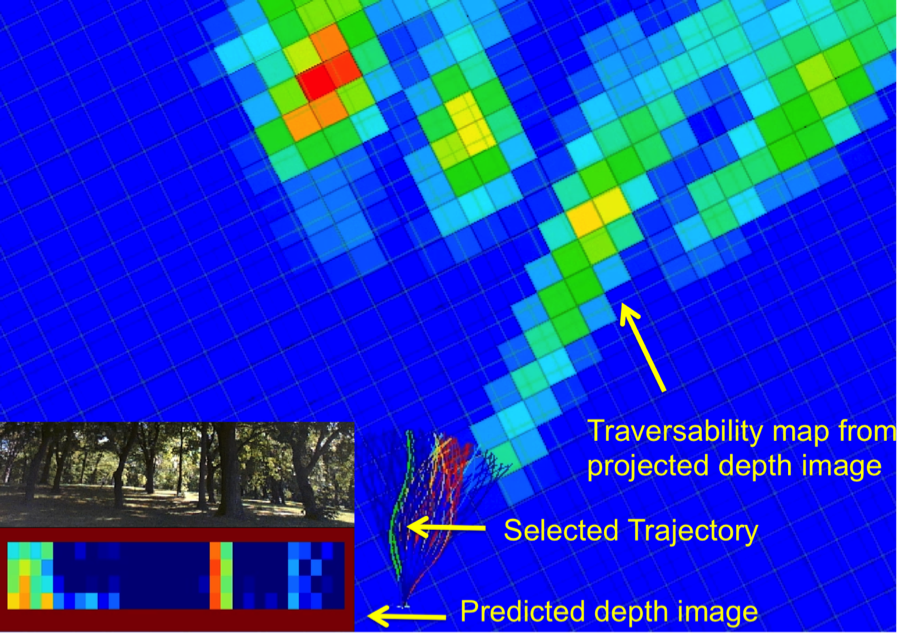}
   \caption{Example of receding horizon with a quadrotor using monocular vision. The lower left images show the view from the front camera and the corresponding depth images from the monocular depth perception layer. The rest of the figure shows the overhead view of the quadrotor and the traversability map (built by projecting out the depth image) where red indicates higher obstacle density. The grid is $1x1$ $m^2$. The trajectories are evaluated on the projected depth image and the one with the least collision score (thick green) trajectory followed.}
   \label{drone_outdoors}
\end{figure}

We demonstrate the \emph{first} receding horizon control with monocular vision implementation on a UAV. Figure \ref{drone_outdoors} shows our quadrotor evaluating a set of trajectories on the projected depth image obtained from monocular depth prediction and traversing the chosen one.

This is motivated by our previous work \cite{ross2013learning} we used imitation learning to learn a purely reactive controller for flying a UAV using only monocular vision through dense clutter. While good obstacle avoidance behavior was obtained, there are certain limitations of a purely reactive layer that a more deliberative approach like receding horizon control can ameliorate. Reactive control is by definition myopic, i.e. it concerns itself with avoiding the obstacles closest to the vehicle. This can lead to it being easily stuck in cul-de-sacs. Since receding horizon control plans for longer horizons it achieves better plans and minimizes the chances of getting stuck \cite{knepper2009diversity}. Another limitation of pure reactive control is the difficulty to reach a goal location or direction. In a receding horizon control scheme, trajectories are selected based on a score which is the sum of two terms: first, the collision score of traversing it and second, the heuristic cost of reaching the goal from the end of the trajectory. By weighting both these terms suitably, goal-directed behavior is realized while maintaining obstacle-avoidance capability. But it is to be noted that reactive control can be integrated with receding horizon for obtaining the best of both worlds in terms of collision avoidance behavior.

Receding horizon control needs three working components 

\begin{itemize}

 \item \emph{A method to estimate depth}: This can be obtained from stereo vision \cite{schmid2014autonomous,matthies2014stereo} or dense structure-from-motion (SfM) \cite{wendel2012dense}. But these are not amenable for achieving higher speeds due to high computational expense. We note that in the presence of enough computation power, information from these techniques can be combined with monocular vision to improve overall perception. 

 Biologists have found strong evidence that birds and insects use optical flow to navigate through dense clutter \cite{srinivasan2011visual}. Optical flow has been used for autonomous flight of UAVs \cite{beyeler2009vision}. However, it tends to be very noisy and therefore difficult to directly derive a robust control principle from. Therefore we follow the same data driven principle as our previous work \cite{ross2013learning} and use local statistics of optical flow to serve as features in the monocular depth prediction module. This allows the learning algorithm to derive complex behaviors in a data driven fashion.

\item \emph{A method for relative pose estimation}: In order to track the chosen trajectory at every cycle, it is needed to estimate the relative state of the vehicle for the duration of the tracking. We developed and demonstrate a relative pose estimation system using a downward facing camera and a sonar to estimate height (Section \ref{pose_estimation})

 \item \emph{A method to deal with perception uncertainty}: Most planning schemes either assume that perception is perfect or make simplistic assumptions of uncertainty. We introduce the concept of making multiple, relevant yet diverse predictions for incorporating perception uncertainty into planning. The intuition is predicated on the observation that avoiding a small number of ghost obstacles is acceptable as far as true obstacles are not missed (high recall, low precision). The details are presented in Section \ref{multiple_predictions}. We demonstrate in experiments the efficacy of this approach as compared to making only a single best prediction.

 \end{itemize}

Our list of contributions in this work are:
\begin{itemize}

\item Budgeted near-optimal feature selection and fast non-linear regression for monocular depth prediction

\item Real time relative vision-based pose estimation

\item Multiple predictions to efficiently incorporate uncertainty in the planning stage.

\item First complete receding horizon control implementation on a UAV.

\end{itemize}

Section \ref{hardware_and_software} describes our hardware and software setup, while Section \ref{multiple_predictions} and Section \ref{depth_prediction} describe the multiple predictions approach to handling uncertainty, and the budgeted depth prediction pipeline. The high frame-rate optical flow based pose estimation system developed for short-term trajectory tracking is described in Section \ref{pose_estimation}. Section \ref{planning_and_control} describes the planning and control framework in detail, and Section \ref{experiments} describes the outdoor experiments and results.

\section{Approach}
\label{approach}

Developing and testing all the integrated modules of receding horizon is very challenging. Therefore we developed a testing protocol where we assembled a rover in addition to a UAV to be able to test various modules separately. Here we describe the hardware platforms and software architecture of our system.

\subsection{Hardware and Software Overview}
\label{hardware_and_software}
In this section we describe the hardware platforms used in our experiments.

\begin{figure}[htbp!]
  \centering
    \includegraphics[width=0.45\textwidth]{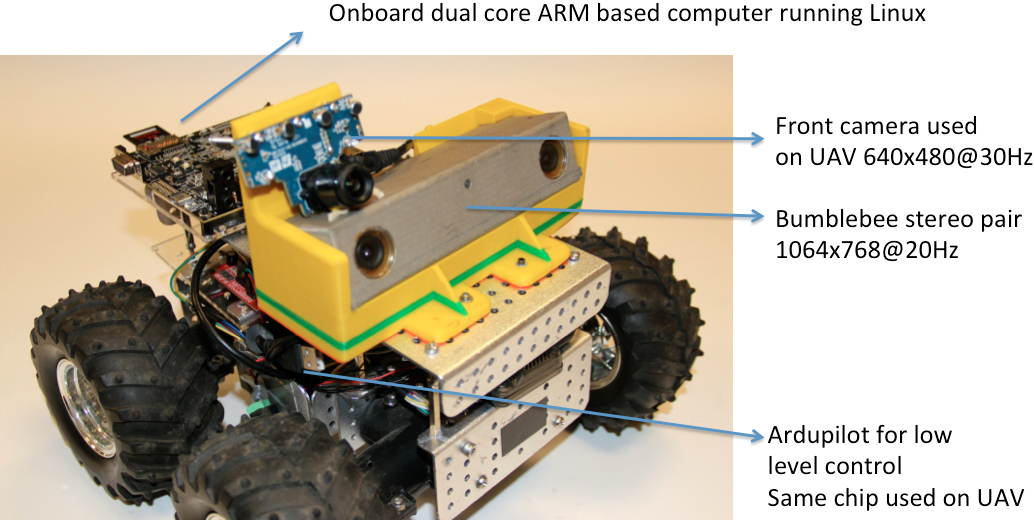}
   \caption{Rover assembled with the same control chips and perception software as UAV for rapid tandem development and validation of modules.}
   \label{rover_glam_shot}
\end{figure}

In order to facilitate parallel testing of perception and planning modules while the UAV control systems were being developed, we assembled a ground rover (Figure \ref{rover_glam_shot}) using off-the-shelf parts.

\subsubsection{Rover}
The rover (Figure \ref{rover_glam_shot}) uses an Ardupilot micrcontroller board \cite{ardupilot} which takes in high level control commands from the planner and controls four motors to achieve the desired motion. The rover is skid-steered. Each motor also has integrated wheel encoders at $1633$ counts per revolution. In order to get short-term relative pose for following a trajectory in receding-horizon control we developed an odometry package which runs on the Ardupilot.

Other than the low-level controllers all other aspects of the rover are kept exactly the same as the UAV to allow seamless transfer of software. For example
the rover has a front facing PlayStation Eye camera which is also used as the front facing camera on the UAV. 

A Bumblebee color stereo camera pair ($1024\times768$ at $20$ fps) is rigidly mounted with respect to the front camera using a custom $3$D printed fiber plastic encasing. This is used for collecting data with groundtruth depth values (Section \ref{depth_prediction}) and validation of planning (Section \ref{planning_and_control}). We calibrate the rigid body transform between the front camera and the left camera of the stereo pair. Stereo depth images and front camera images of the environment are recorded simultaneously while driving the rover around using a joystick. The depth images are then transformed to the front camera's coordinate system to provide groundtruth depth values for every pixel.

\subsubsection{UAV}
\begin{figure}[htbp!]
  \centering
    \includegraphics[width=0.45\textwidth]{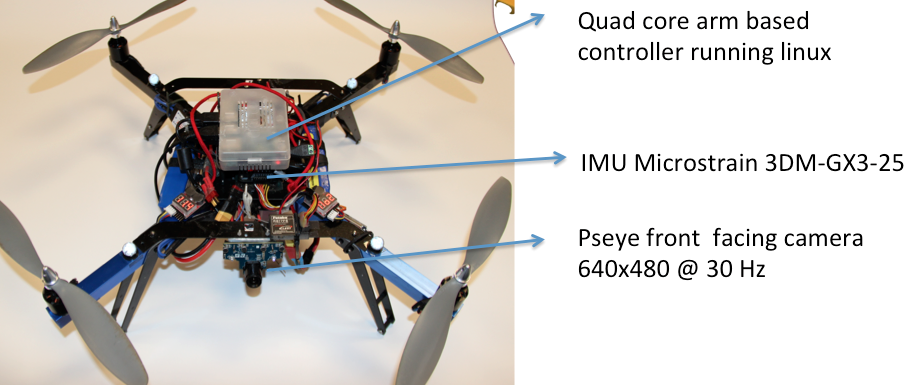}
   \caption{Quadrotor used as our development platform.}
   \label{uav_glam_shot}
\end{figure}

\begin{figure}[htbp!]
  \centering
    \includegraphics[width=0.45\textwidth]{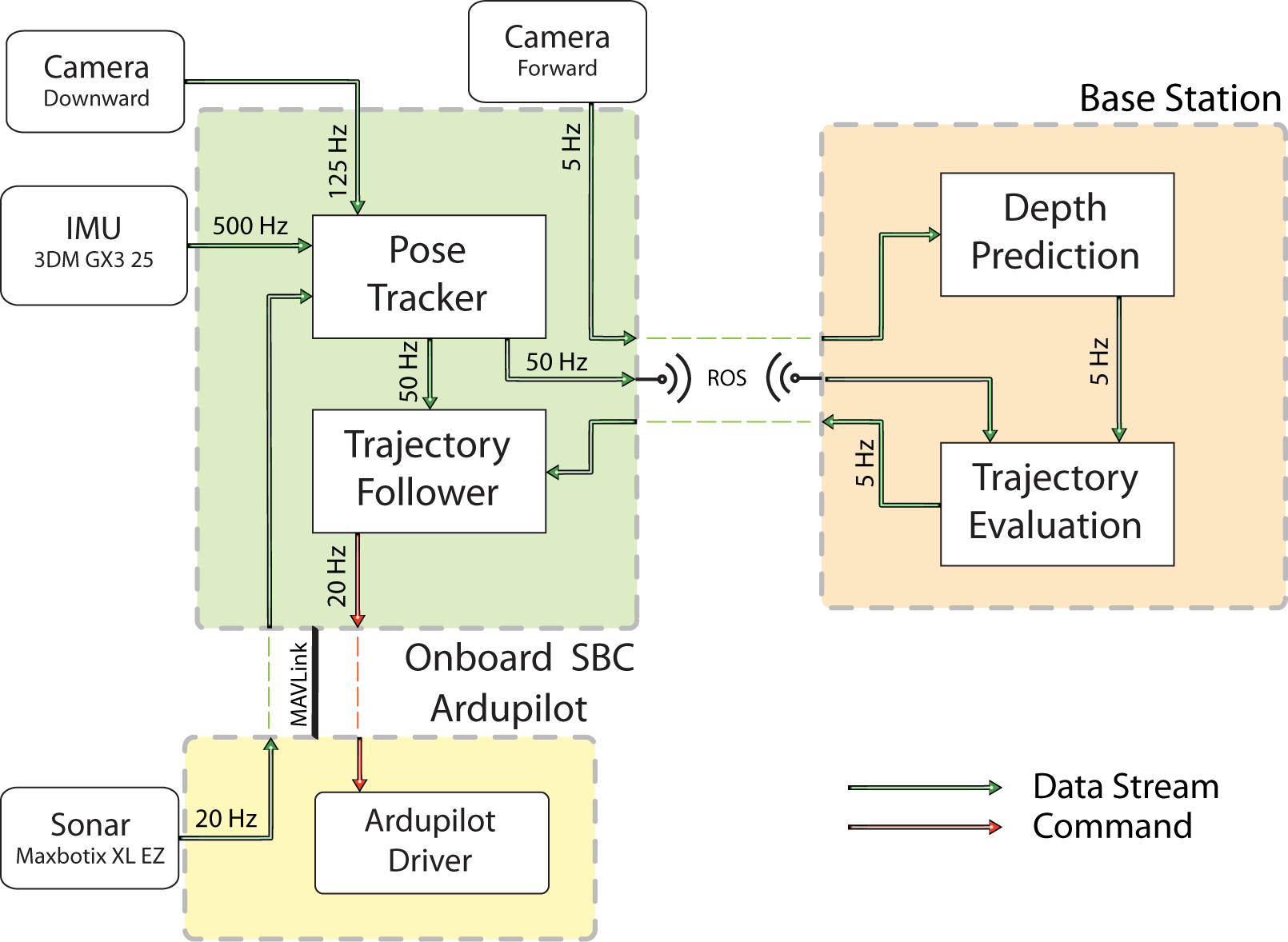}
   \caption{Schematic diagram of hardware and software modules}
   \label{bird_architecture}
\end{figure}

Figure \ref{uav_glam_shot} shows the quadrotor we use for our experiments. Figure \ref{bird_architecture} shows the schematic of the various modules that run onboard and offboard. The base chassis, motors and autopilot are assembled using the Arducopter kit \cite{ardupilot}. Due to the excessive drift and noise of the IMU integrated in the Ardupilot unit, we added a Microstrain 3DM GX3 25 IMU which is used to aid real time pose estimation. There are two PlayStation Eye cameras: one facing downwards for real time pose estimation, one facing forward. The onboard processor is a quad-core ARM based computer which runs Ubuntu and ROS \cite{ros2009}. 
 This unit runs the pose tracking and trajectory following modules. A sonar is used to estimate altitude. The image stream from the front facing camera is streamed to the base station where the depth prediction module processes it; the trajectory evaluation module then finds the best trajectory to follow to minimize probability of collision and transmits it to the onboard computer where the trajectory following module runs a pure pursuit controller to do trajectory tracking \cite{coulter1992implementation}. The resulting high level control commands are sent to the Ardupilot which sends low level control commands to the motor controllers to achieve the desired motion.

\subsection{Monocular Depth Prediction}
\label{depth_prediction}
In this section we describe the depth prediction approach from monocular images, budgeted feature selection and the fast non-linear regression method used for regression. 
 
\begin{figure}[htbp!]
  \centering
    \includegraphics[width=0.2\textwidth]{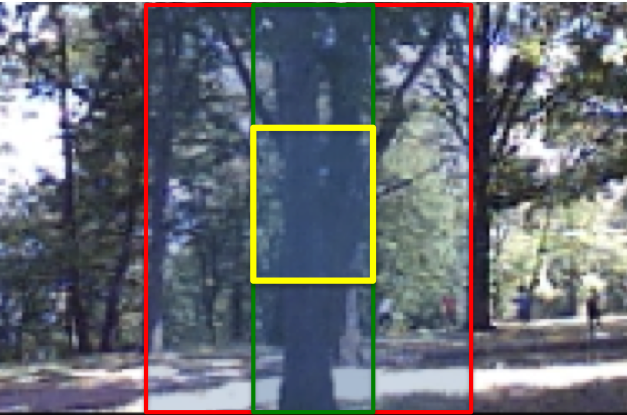}
   \caption{The yellow box is an example patch, the green box is the column of the same width surrounding it, and the red box is the column of $3$ times the patch width surrounding it. Features are extracted individually at the patch, and the columns and concatenated together to form the total feature representation of the patch.}
   \label{patch_column}
\end{figure}

In image is first gridded up into non-overlapping patches. We predict the depth in meters at every patch of the image (Figure \ref{patch_column} yellow box). For each patch we extract features which describe the patch, features which describe the full column containing the patch (Figure \ref{patch_column} green box) and features which describe the column of three times the patch width (Figure \ref{patch_column} red box), centered around the patch. The final feature vector for a patch is the concatenation of the feature vectors of all three regions. When a patch is seen by itself it is very hard to tell the relative depth with respect to the rest of the scene. But by adding the features of the surrounding area of the patch, more context is available to aid the predictor. 

\subsubsection{Description of features}
In this part we describe in brief the features used to represent the patch. We use the features as used in previous work on monocular imitation learning \cite{ross2013learning} for UAVs, which are partly inspired by the work of Saxena et al., \cite{saxena2005learning}. In this case, instead of regressing directly from features of an image to a control action, we predict the depth at every patch which is then used by the planning module.

\begin{itemize}

\item \emph{Optical flow}: We use the Farneback dense optical flow \cite{farneback2003two}  implementation in OpenCV to compute for every patch the average, minimum and maximum optical flow values which are used as feature descriptors of that patch. 

\item \emph{Radon Transform}: The radon transform captures strong edges in a patch \cite{helgason1980support}.

\item \emph{Structure Tensor}:
The structure tensor describes the local texture of a patch \cite{harris1998combined}.
 
\item \emph{Laws' Masks}: These describe the texture intensities \cite{davies2004machine}. For details on radon transform, structure tensor and Laws' masks usage see \cite{ross2013learning}.

\item \emph{Histogram of Oriented Gradients (HoG)}: This feature has been used widely in the computer vision community for capturing texture information for human pose estimation as well as object detection \cite{dalal2006human}. For each patch we compute the HoG feature in $9$ orientation bins. We subdivide the patch into cells such that each resulting cell's side is at least half the width of the patch.

\item \emph{Tree feature}: We use the per pixel fast classifier by Li et al. \cite{li2013pixel} to train a supervised tree detector. Li et al. originally used this for real time hand detection in ego-centric videos. For a given image patch we use this predictor to output the probability of each pixel being a tree. This information is then used as a feature for that patch.

\end{itemize}

\subsubsection{Data Collection}
RGB-D sensors like the Kinect, currently do not work outdoors. Since camera and calibrated nodding lidar setup is expensive and complicated we used a rigidly mounted Bumblebee stereo color camera and the PlayStation Eye camera for our outdoor data collection. This setup was mounted on the rover as shown in Figure \ref{rover_glam_shot}. We have collected data at $4$ different locations with tree density varying from low to high, under varying illumination conditions and in both summer and winter conditions. Our corpus of imagery with stereo depth information is around $16000$ images and growing. We plan on making this dataset publicly available in the near future.

\subsubsection{Fast Non-linear Prediction}
Due to harsh real-time constraints an accurate but fast predictor is needed. Recent linear regression implementations are very fast and can operate on millions of features in real time \cite{vw} but are limited in predictive performance by the inherent linearity assumption. In very recent work Agarwal et al. \cite{agarwal2013least} develop fast iterative methods which use linear regression in the inner loop to obtain overall non-linear behavior. This leads to fast prediction times while obtaining much better accuracy. We implemented Algorithm $2$ in \cite{agarwal2013least} and found that it lowered the error by $10$ \% compared to just linear regression, while still allowing real time prediction.

\subsection{Budgeted Feature Selection}
While there are many kinds of visual features that can be extracted, they need to be computed in real time. The faster the desired speed of the vehicle, the faster the perception and planning modules have to work to maintain safety. Compounded with the limited computational power available onboard a small UAV, this imposes a very challenging budget within which to make a prediction. Each kind of feature requires different time periods to extract, while contributing different amounts to the prediction accuracy. For example, radon transforms might take relatively less time to compute but contribute a lot to the prediction accuracy, while another feature might take more time but also contribute relatively less or vice versa. This problem is further complicated by the ``grouping'' effects where a particular feature's performance is affected by the presence or absence of other features.

Given a time budget, the naive but obvious solution is to enumerate all possible combinations of features which respect the budget and find the group of features which achieve the minimum loss. This is exponential in the number of available features. Instead we use the efficient approach developed by Hu et al. \cite{hu2014efficient} to efficiently select the near-optimal set of features which meet the imposed budget constraints. Their approach uses a simple greedy algorithm that first whitens feature groups and then recursively chooses groups by the reduction in explained variance divided by the time to achieve that reduction. A more efficient variant of this with equivalent guarantees, chooses features by computing gradients to approximate the reduction in explained variance, eliminating the need to ``try'' all feature groups sequentially. For each specified time budget, the features selected by this procedure are within a constant factor of the optimal set of features which respect that time budget. Since this holds across all time budgets, this procedure provides a recursive way to generate feature sets across time steps.

\begin{figure}[htbp!]
  \centering
    \includegraphics[width=0.4\textwidth]{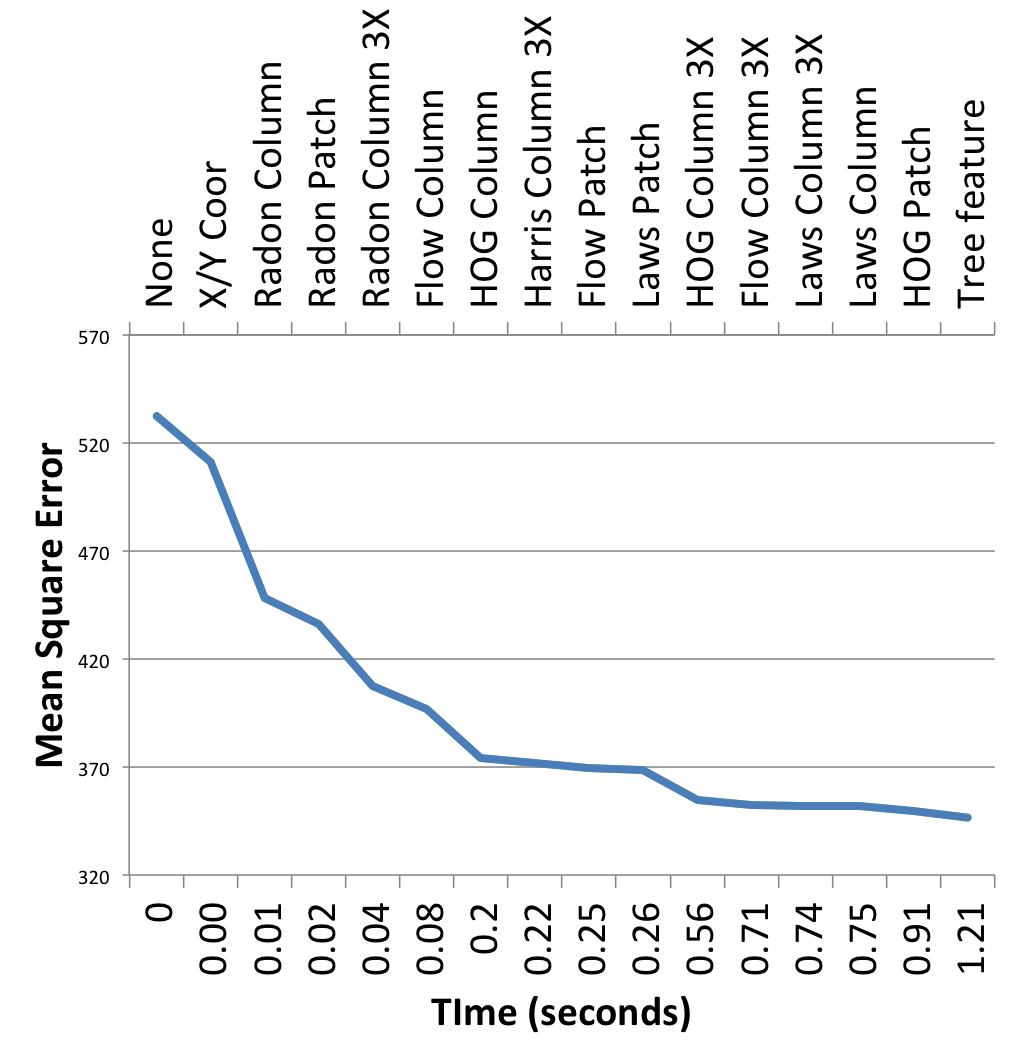}
   \caption{On the upper x-axis the sequence of features selected by Hu et al.'s method \cite{hu2014efficient} and the lower x-axis shows the cumulative time taken for all features up to that point. The near-optimal sequence of features rapidly decrease the prediction error. For a given time budget, the sequence of features to the left of that time should be used.}
   \label{speedboost_plot}
\end{figure}

Figure \ref{speedboost_plot} shows the sequence of features that was selected by Hu et al.'s \cite{hu2014efficient} feature selection procedure. For any given budget only the features on the left up to the specified time budget need to be computed.

\begin{figure}[htbp!]
  \centering
    \includegraphics[width=0.45\textwidth]{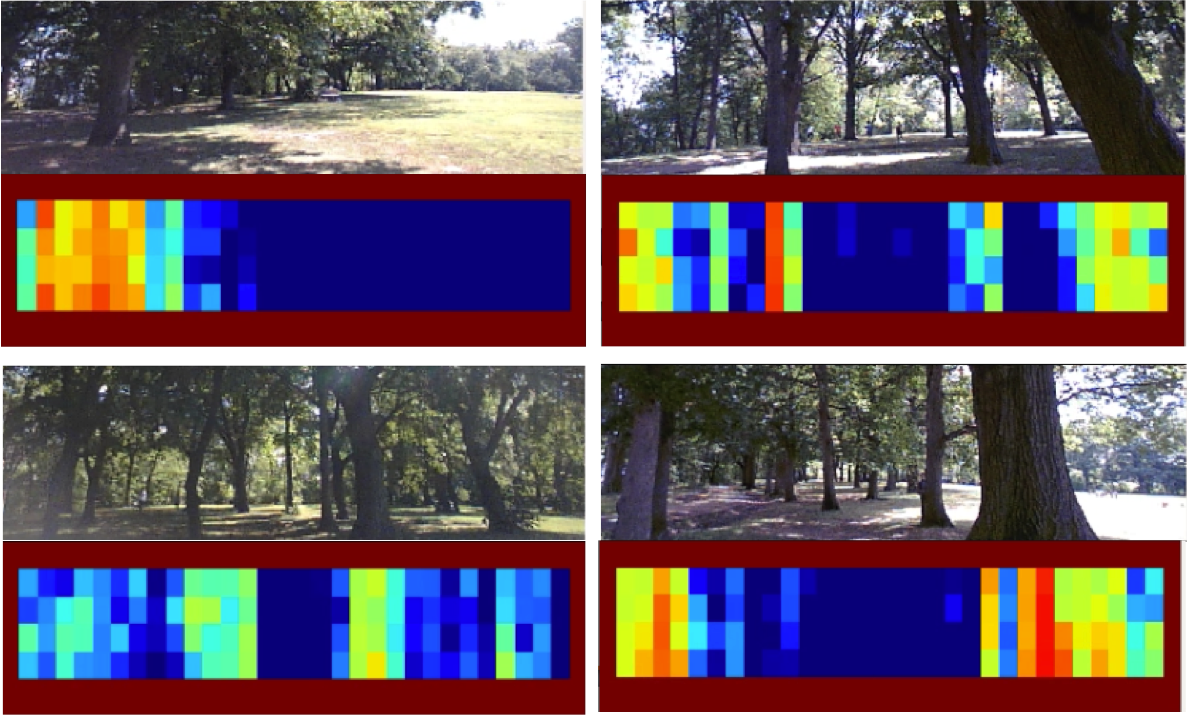}
   \caption{Depth prediction examples on real outdoor scenes. Closer obstacles are indicated by red.}
   \label{depth_examples}
\end{figure}

\subsection{Multiple Predictions}
\label{multiple_predictions}
\begin{figure}[htbp!]
  \centering
    \includegraphics[width=0.5\textwidth]{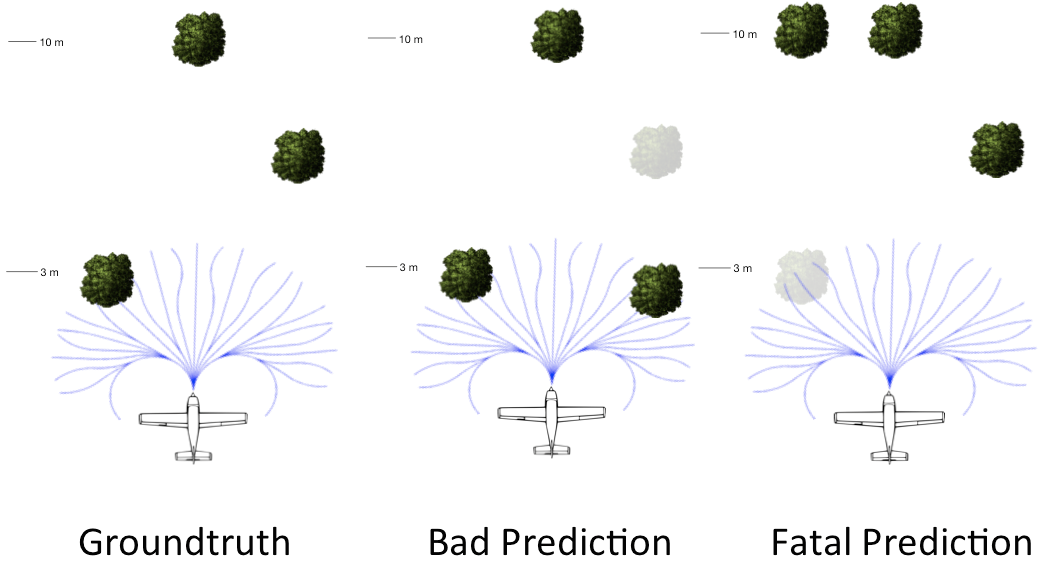}
   \caption{Illustration of the complicated nature of the loss function for collision avoidance. (Left) Groundtruth tree locations (Middle) Bad prediction where a tree is predicted closer than it actually is located (Right) Fatal prediction where a tree close by is mis-predicted further away.}
   \label{hardness_cost_illus}
\end{figure}

The monocular depth estimates are often noisy and often inaccurate due the inherent challenging nature of the problem. A planning system must incorporate this uncertainty to achieve safe flight. Figure \ref{hardness_cost_illus} illustrates the difficulty of trying to train a predictive method for building a perception system for any general collision avoidance problem. Figure \ref{hardness_cost_illus} (left) shows a ground truth location of trees in the vicinity of an autonomous UAV flying through the forest. Figure \ref{hardness_cost_illus} (middle) shows the location of the trees as predicted by the onboard perception system. In this prediction the trees on the left and far away in front are predicted correctly but the tree on the right is predicted close to the UAV. This will cause the UAV to dodge a ghost obstacle. While this is bad it is not fatal because the UAV will not crash but just have to make some spurious motions. But the prediction of trees in Figure \ref{hardness_cost_illus} (right) is potentially fatal. Here the trees far away in front and on the right are correctly predicted where as the tree on the left originally close to the UAV is mis-predicted to be far away. This type of mistake will cause the UAV to crash into an obstacle it does not know is there. 

Ideally, a vision-based perception system should be trained to minimize loss functions which will penalize such fatal predictions more than other kind of predictions. But even writing down such a loss function is difficult. Therefore most monocular depth perception systems try to minimize easy to optimize surrogate loss functions like regularized $L_1$ or $L_2$ loss \cite{saxena2005learning}. We try to reduce the probability of collision by generating multiple interpretations of the scene in front of the UAV to hedge against the risk of committing to a single interpretation of the scene which could be potentially fatal as illustrated above in Figure \ref{hardness_cost_illus}. We generate $3$ interpretations of the scene and evaluate the trajectories in all $3$ interpretations simultaneously. The trajectory which is least likely to collide on average in all interpretations is then chosen as the trajectory to traverse.

One way of making multiple predictions is to just sample the posterior distribution of a learnt predictor. In order to truly capture the uncertainty of the predictor, a lot of interpretations have to be sampled and trajectories evaluated on each of them. A large number of samples will be from around the peaks of this distribution leading to wasted samples. This is not feasible given the real time constraints of the problem.

In previous work \cite{deyConSeqOptRSS}, we have developed techniques for predicting a budgeted number of interpretations of an environment with applications to  manipulation, planning and control. Batra et al., \cite{batra2012diverse} have also applied similar ideas to structured prediction problems in computer vision. These approaches try to come up with a small number of relevant but diverse interpretations of the scene so that at least one of them is correct. In this work, we adopt a similar philosophy and use the error profile of the fast non-linear regressor described in Section \ref{depth_prediction} to make two additional predictions: The non-linear regressor is first trained on a dataset of $14500$ images and it's performance on a held-out dataset of $1500$ images is evaluated. For each depth value predicted by it, the average over-prediction and under-prediction error is recorded. For example the predictor may say that an image patch is at $3$ meters while on average whenever it says so, it is actually either, on average, at $4$ meters or at $2.5$ meters. We round each prediction depth to the nearest integer, and record the average over and under-predictions as in the above example in a look-up table (LUT). At test time the predictor produces a depth map and the LUT is applied to this depth map, producing two additional depth maps: one for over-prediction error, and one for the under-prediction error.

\begin{figure}[htbp!]
  \centering
    \includegraphics[width=0.4\textwidth]{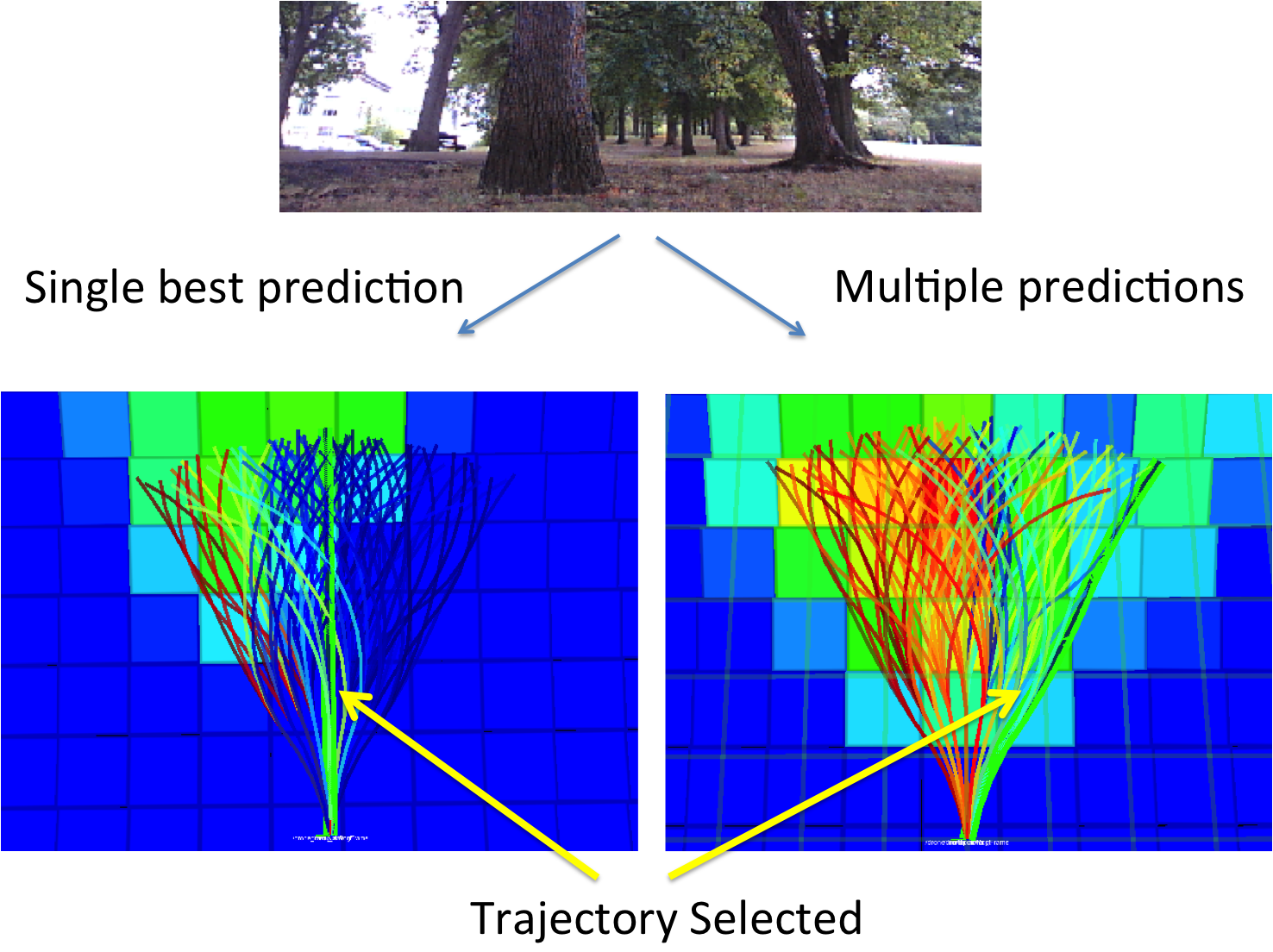}
   \caption{The scene at top is an example from the front camera of the UAV. On the left is shown the predicted traversability map (red is high cost, blue is low cost) resulting from a single interpretation of the scene. Here the UAV has selected the straight path (thick, green) which will make it collide with the tree right in front. While on the right the traversability map is constructed from multiple interpretations of the image, leading to the trajectory in the right being selected which will make the UAV avoid collision.}
   \label{multiple_world_advantage}
\end{figure}

Figure \ref{multiple_world_advantage} shows an example in which making multiple predictions is clearly beneficial compared to the single best interpretation. We provide more experimental details and statistics in Section \ref{experiments}.

\subsection{Pose Estimation}
\label{pose_estimation}
As discussed before, a relative pose-estimation system is needed to follow the trajectories chosen by the planning layer. We use a downward looking camera in conjunction with a sonar for determining relative pose. Looking forward to determine pose is ill-conditioned due to a lack of parallax as the camera faces the direction of motion. There are still significant challenges involved when looking down. Texture is often very self similar making it challenging for traditional feature based methods to be employed.

Dense methods such as DTAM \cite{newcombe2011dtam} build a complete 3D model of the world that make tracking very easy, but require GPUs to run in real time. As a consequence, these can not be deployed onboard our platform. The other alternative is to use sparse mapping approaches like \cite{klein2007parallel} that extract robustly identifiable interest points that are tracked over consecutive frames using a wide variety of feature descriptors, but fail due to issues mentioned above.

In receding horizon we do not need absolute pose with respect to some fixed world coordinate system. This is because we need to follow trajectories for short durations only. So as long as we have a relative, consistent pose estimation system for this duration ($3$ seconds in our implementation) we can successfully follow trajectories. Thus by using only differential motion between consecutive frames and not consciously refining pose information, we can get away with small drift over time.

This motivated the use of a variant of a simple algorithm that has been presented quite often, most recently in \cite{honegger2013open}. This approach involves using a Kanade-Lucas-Tomasi (KLT) tracker \cite{tomasi1991detection} to detect where each pixel in a grid of pixels moves over consecutive frames, and estimating the mean flow from these after rejecting outliers. The KLT tracker operates on the principle of assuming a small consistent flow from a patch around a feature point in the first image to another in the second image. It does this by performing iterative Newton-Raphson refinement steps minimizing the error in image intensities for each pixel over the patch weighted by the local second derivative around the interest point. This estimate of flow however tries to find the best planar displacement between the two patches, and does not take into account out-of-plane rotations, for instance, due to motion of the camera. Camera ego-motion is compensated using motion information from the IMU. Finally the scene depth is estimated from a sonar. We obtain instantaneous relative velocity between the camera and ground which we integrate over time to get position. 

This process is computationally inexpensive, and can thus be run at very high frame rates. Higher frame rates lead to smaller displacements between pairs of images, which in turn makes tracking easier to compute and more accurate since over short displacements first order approximations to nonlinearities have less error, thus driving down the overall error in the system. 

\begin{figure}[t]
  \centering
    \includegraphics[width=0.45\textwidth]{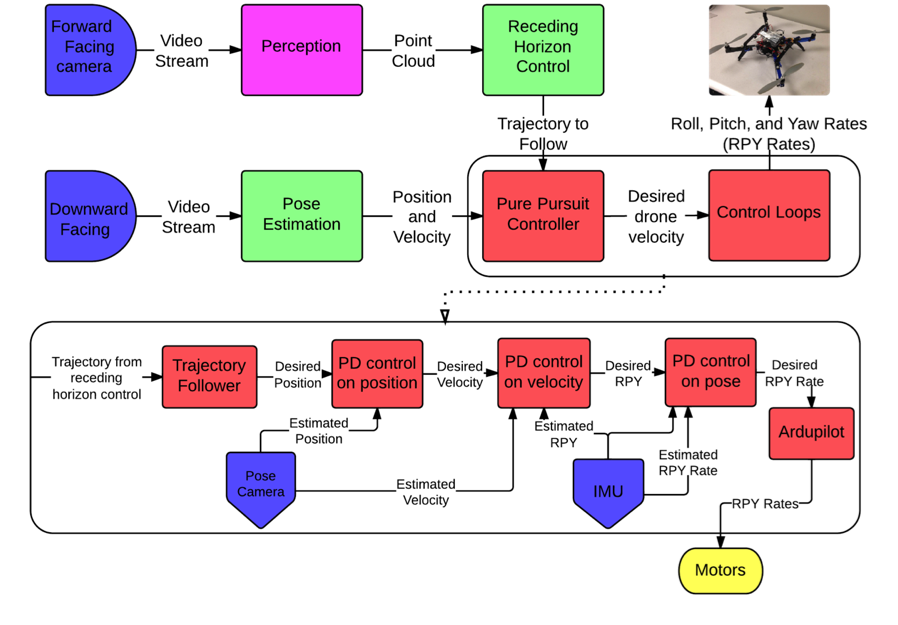}
   \caption{The overall flow of data and control commands between various modules. The pure pursuit trajectory follower and low level control loops (red boxes) are shown in greater detail at the bottom.}
   \label{flow_of_commands}
\end{figure}

\subsubsection{Implementation Details}
We obtain the angular velocity of the camera using the output of the IMU driver that provides us orientation and velocity updates at upto 500Hz. The optical flow of the central patch is determined by using OpenCV's pyramidal Lucas-Kanade tracker implementation.

For robustness to outliers, we evaluate flow for a small discrete grid around the central patch and take the mean of the unrotated optical flow vectors as the final flow vector. We also do an outlier detection step by comparing the standard deviation of the flow vectors obtained for every pixel on the grid to a specific threshold. Whenever the variance of the flow is high, we do not calculate the mean flow velocity, and instead decay the previous velocity estimate by a constant factor. However, such situations rarely occur due to the very high frame rate of the camera feed we use that ensures that only a small change in the flow field occurs between consecutive images.
 

\subsubsection{Performance vs Ground Truth}
\begin{figure}[htbp!]
\centering
\begin{tabular}{cc}
\includegraphics[width=0.45\linewidth]{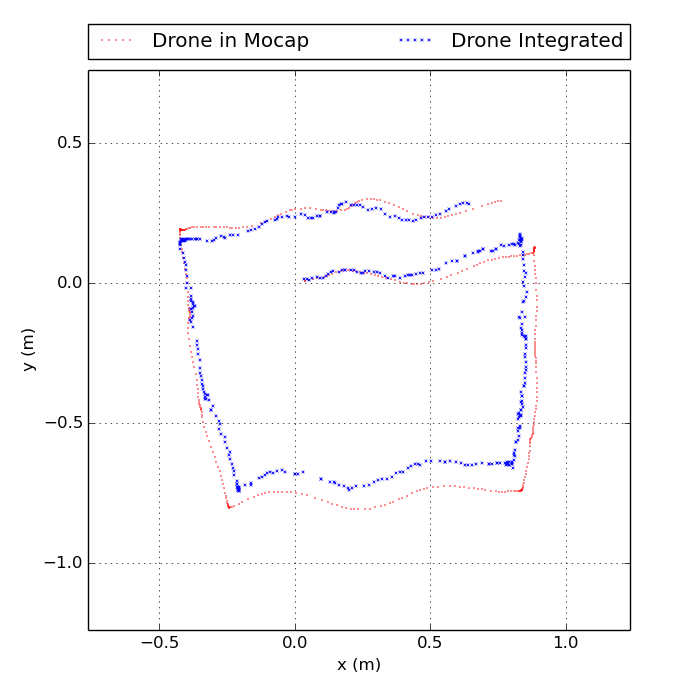}  & \includegraphics[width=0.45\linewidth]{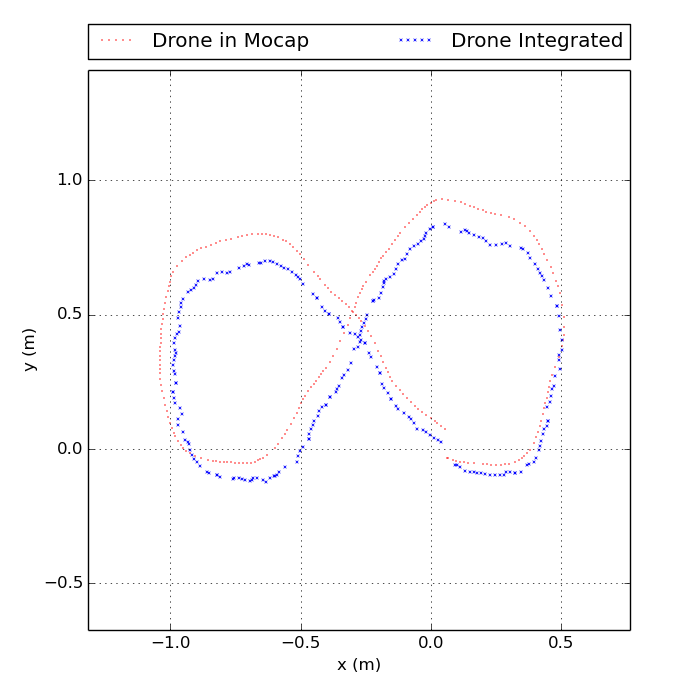}
\end{tabular}
\caption{Comparison of the differential flow tracker performance vs ground truth in MOCAP. Red tracks are the trajectories in MOCAP, blue are those determined by the algorithm. Note that the formulation of the receding horizon setup is such that mistakes made in following a specific trajectory are forgiven up to an extent since we replan every few seconds.}
\label{fig:tracking}
\end{figure}

To evaluate the performance of the differential flow based tracker, we took the camera-IMU quadrotor setup to the motion capture lab and compared the tracking performance. We held the quadrotor in our hand and walked over artificial camouflage texture. The resulting tracks are as shown in figure \ref{fig:tracking}

\begin{figure}[tbp!]
    \centering
	\begin{tabular}{cc}
	            \raisebox{.25\height}{ \includegraphics[width=0.3\linewidth, trim=100 75 100 75, clip=true]{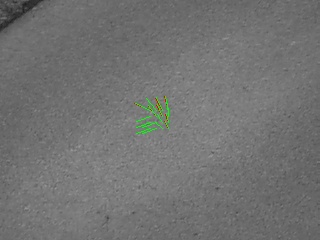}} & \raisebox{.25\height}{\includegraphics[width=0.3\linewidth, trim=100 75 100 75, clip=true]{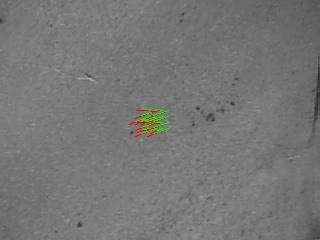}}
    \end{tabular}
    \vspace{-7mm}
	\caption{Instances of failure of the pose tracking system over challenging surfaces. Note the absence of texture in these 320x240 images. The figure shows the flow tracks corresponding to the points on the grid. Red tracks show the uncorrected optical flow, while the green tracks (superimposed) show the flow vectors `unrotated' using the IMU.}
	\label{fig:flow_visualization}
\end{figure}
%
%

\subsection{Planning and Control}
\label{planning_and_control}
Figure \ref{flow_of_commands} shows the overall flow of data and control commands in the architecture. The front facing camera video stream is fed to the perception module which predicts the depth of every pixel in a frame, projects it to a point cloud representation and sends it to the receding horizon control module. A trajectory library of $78$ trajectories of length $5$ meters is budgeted and picked from a much larger library of $2401$ trajectories using the maximum dispersion algorithm by Green et al. \cite{green2006paths}. This is a greedy procedure for selecting trajectories, one at a time, so that each subsequent trajectory spans maximum area between it and the rest of the trajectories. The receding horizon module maintains a score value for every point in the point cloud. The score of a point decays exponentially the longer it exists. After some time when it drops below a user set threshold the point is deleted. The rate of decay is specified by setting the time constant of the decaying function. This fading memory representation of the local scene layout has two advantages: 1) It prevents collisions caused by narrow field-of-view issues where the quadrotor forgets that it has just avoided a tree, sees the next tree and dodges sideways, crashing into the just avoided tree. 2) It allows emergency backtracking maneuvers to be safely executed if required, since there is some local memory of the obstacles it has just passed.

Our system can accept any goal direction as input and ensures that the vehicle makes progress towards the goal while avoiding obstacles along the way.
The score for each trajectory is the sum of three terms: 1) A sphere of the same radius as the quadrotor is convolved along a trajectory and the score of each point in collision is added up. The higher this term relative to other trajectories, the higher the likelihood of this trajectory being in collision. 2) A term which penalizes a trajectory whose end \emph{direction} deviates from goal direction. This is weighted by a user specified parameter. This term induces goal directed behavior and is tuned to ensure that the planner always avoids obstacles as a first priority. 3) A term which penalizes a trajectory for deviating in \emph{translation} from the goal direction.

The pure pursuit controller module (Figure \ref{flow_of_commands}) takes in the coordinates of the trajectory to follow and the current pose of the vehicle from the optical flow-based pose estimation system (Section \ref{pose_estimation}). We use a pure pursuit strategy \cite{coulter1992implementation} to successfully track it. Specifically, this involves finding the closest point on the trajectory from the robot's current estimated position and setting the target waypoint to be a certain fixed lookahead distance further along the trajectory. The lookahead distance can be tuned to obtain the desired smoothness while following the trajectory; concretely, a larger lookahead distance leads to smoother motions, at the cost of not following the trajectory exactly. Using the pose updates provided by the pose estimation module, we head towards this moving waypoint using a generic PD controller. Since the receding horizon control module continuously replans (at $5$ hz) based on the image data provided by the front facing camera, we can choose to follow arbitrary lengths along a particular trajectory before switching over to the latest chosen one.

\subsubsection{Validation of Modules}
We validated each module separately as well as in tandem with other modules where each validation was progressively integrated with other modules. This helped reveal bugs and instabilities in the system.

\begin{figure}[htpb!]
  \centering
    \includegraphics[width=0.3\textwidth]{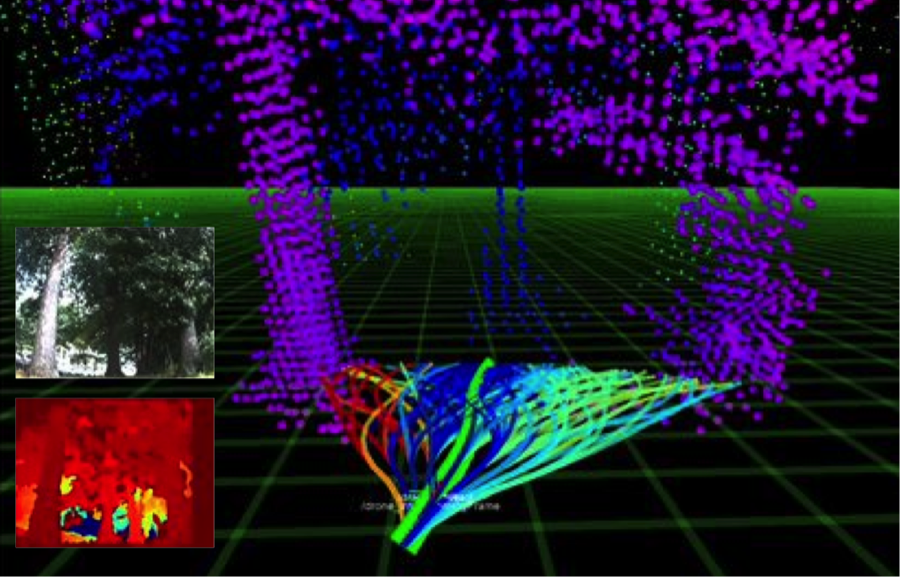}
   \caption{Receding horizon control validation with rover using depth images from stereo. The bright green trajectory is the currently selected trajectory to follow. Red trajectories indicate that they are more likely to be in collision.}
   \label{rover_stereo}
\end{figure}

\begin{figure}[htpb!]
  \centering
    \includegraphics[width=0.3\textwidth]{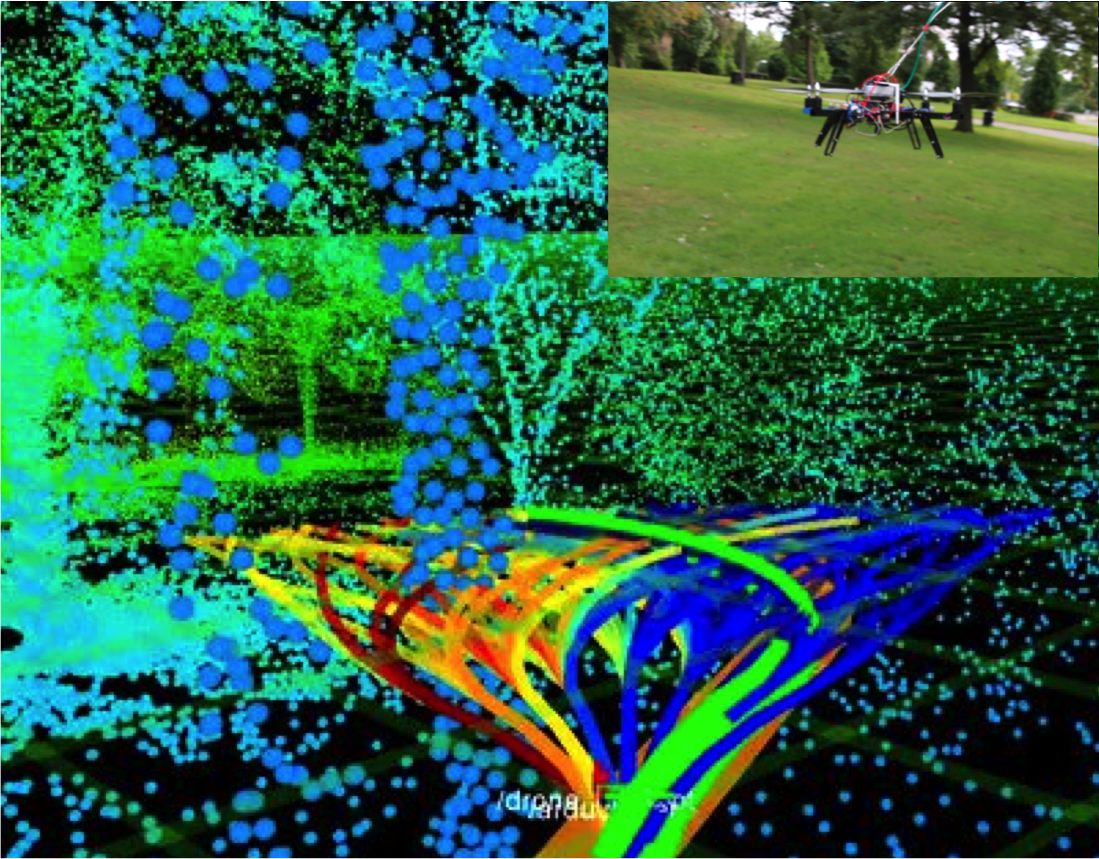}
   \caption{Hardware-in-the-loop testing with UAV in open field. The receding horizon module was fooled into thinking that it was in the midst of a real world point cloud while it planned and executed its way through it. This allowed us to validate planning and control without endangering the UAV.}
   \label{drone_hwil}
\end{figure}

\begin{itemize}

\item \emph{Trajectory Evaluation and Pure Pursuit Validation with Stereo Data on Rover}: We tested the trajectory evaluation and pure pursuit control module by running the entire pipeline (other than monocular depth prediction) with stereo depth images on the rover. Figure \ref{rover_stereo}.

\item \emph{Trajectory Evaluation and Pure Pursuit Validation with Monocular Depth on Rover}: This test is the same as above but instead of using depth images from stereo we used the perception system's depth prediction. This allowed us to tune the parameters for scoring trajectories in the receding horizon module to head towards goal without colliding with obstacles.

\item \emph{Trajectory Evaluation and Pure Pursuit Validation with Known Obstacles in Motion Capture on UAV}: While testing of modules progressed on the rover we assembled and developed the pose estimation module (Section \ref{pose_estimation}) for the UAV. We tested this module in a motion capture lab where the position of the UAV as well of the obstacles was known and updated at $120$ Hz. (See Figure \ref{rec_horz_illus_mocap})

\item \emph{Trajectory Evaluation and Pure Pursuit Validation with Hardware-in-the-Loop (HWIL)}: In this test we ran the UAV in an open field, fooled the receding horizon module to think it was in the midst of a point cloud and ran the whole system (except perception) to validate planning and control modules. Figure \ref{drone_hwil} shows an example from this setup.

\item \emph{Whole System}: After validating each module following the evaluation protocol described above, we ran the whole system end-to-end. Figure \ref{drone_outdoors} shows an example scene of the quadrotor in full autonomous mode avoiding trees outdoors. We detail the results of collision avoidance in Section \ref{experiments}.

\end{itemize}

\section{Experiments}
\label{experiments}
In this section we analyze the performance of our proposed deliberative approach for autonomous navigation of a UAV in cluttered natural environments using only monocular vision. All the experiments were conducted in a densely cluttered forest area, while restraining the drone through a light-weight tether.

\begin{figure}[!htb]
\centering
\begin{tabular}{c @ {  } c }
\includegraphics[width=0.55\linewidth]{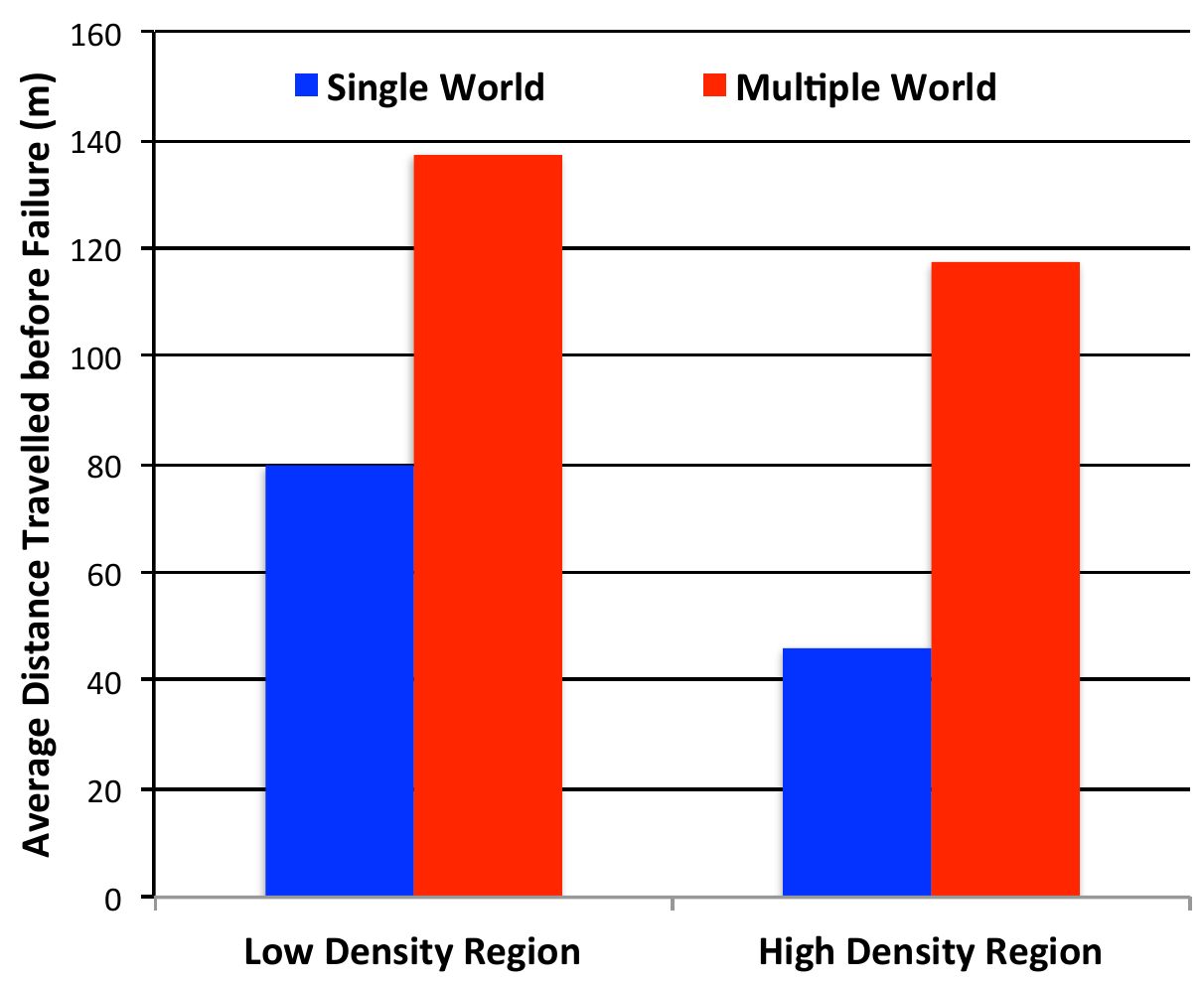} & 
\includegraphics[width=0.35\linewidth]{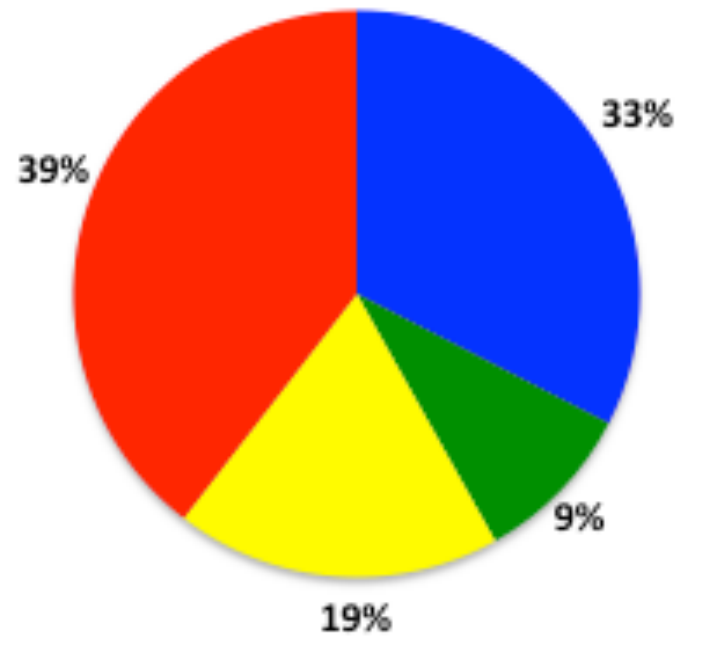}
\end{tabular}
\caption{(a) Average distance flown by the drone before a failure. (b) Percentage of failure for each type. Red: Large Trees, Yellow: Thin Trees, Blue: Foliage, Green: Narrow FOV.} 
\label{fig:result}
\end{figure}

Quantitatively, we evaluate the performance of our system by observing the average distance flown autonomously by the UAV over several runs, before an intervention. An intervention, in this context, is defined as the point at which the pilot needs to overwrite the commands generated by our control system so as to prevent the drone from crashing. Experiments were performed using both our proposed multiple predictions approach and single best prediction, and the corresponding comparison has been shown in Fig. \ref{fig:result}. Both tests were performed in regions of high and low clutter density (approx. 1 tree per $6x6$ $m^2$ and $12x12$ $m^2$, respectively). It can be observed that multiple predictions results in significantly better performance. In particular, the drone was able to fly autonomously without crashing over a $137$ m distance for low density regions. The difference is even higher in case of high-density regions where committing to a single prediction can be even more fatal. 

\begin{figure}[htbp!]
  \centering
    \includegraphics[width=0.4\textwidth]{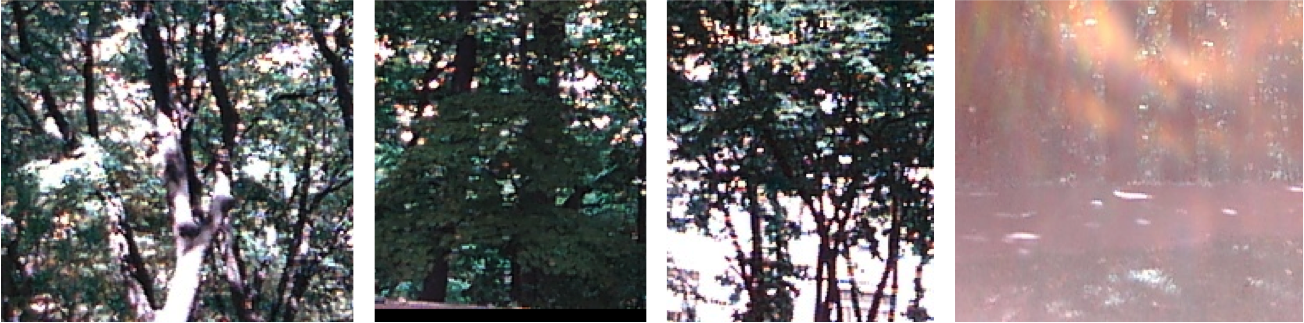}
   \caption{Examples of interventions: (Left) Bright trees saturated by sunlight from behind (Second from left) Thick foliage (Third from left) Thin trees (Right) Flare from direct sunlight. Camera/lens with higher dynamic range and more data of rare classes should improve performance.}
   \label{failure_cases}
\end{figure}

Further, we evaluate the success rate for avoiding the large and small trees using our proposed approach. The results have been tabulated in Table \ref{table:percent}. We are successfully able to avoid $96$ \% of all trees over a total covered distance of more than $1$ km. We extend our evaluation to qualitatively assess, and understand the failure cases responsible for the above results (See Fig. \ref{fig:result}). The type of failures are broken down by the type of obstacle the drone failed to avoid, or whether the obstacle was not in the field-of-view (FOV). Overall, $39$ \% of the failures were due to large trees and $33$ \% on hard to perceive obstacles like branches and leaves. As expected, the narrow FOV is now the least contributor to the failure cases as compared to a more reactive control strategy \cite{ross2013learning}. This is intuitive, since the reactive control is myopic in nature and our deliberate approach helps overcome this problem as described in the previous sections. Figure \ref{failure_cases} shows some typical intervention examples.

\begin{table}[!t]
\begin{center}
\resizebox{0.95\linewidth}{!}{%
\begin{tabular}{ |c|c|c|}
\hline
\rule{0pt}{2ex} {} & \textbf{Multiple Predictions} & \textbf{Single Prediction} \\
\hline
\textbf{Total Distance} {} & 1020 m & 1010 m  \\
\hline
\textbf{Large Trees Avoided}  & 93.1 \% & 84.8 \% \\
\hline
\textbf{Small Trees Avoided} & 98.6 \% & 95.9 \% \\
\hline
\textbf{Overall Accuracy} & \textbf{96.6 \%} & \textbf{92.5 \%} \\
\hline
\end{tabular}}
\end{center}
\caption{Success rate of avoiding trees.}
\label{table:percent}
\end{table}

\section{Conclusion}
\label{conclusion}
While we have obtained promising results from the current approach, a number of challenges remain. For example, better handling of sudden strong wind disturbances and control schemes for better leveraging the full dynamic envelope of the vehicle. In ongoing work we are moving towards complete onboard computing of all modules to reduce latency. We can leverage other sensing modes like sparse, but more accurate depth estimation from stereo, which can be used as ``anchor'' points to improve dense monocular depth estimation. Similarly low power, light weight lidars can be actively foveated to high probability obstacle regions to reduce false positives and get exact depth.

Another central future effort is to integrate the purely reactive \cite{ross2013learning} approach with the deliberative scheme detailed here, for better performance.

\bibliographystyle{IEEEtran}
\bibliography{references}

\begin{thebibliography}{10}
\providecommand{\url}[1]{#1}
\csname url@rmstyle\endcsname
\providecommand{\newblock}{\relax}
\providecommand{\bibinfo}[2]{#2}
\providecommand\BIBentrySTDinterwordspacing{\spaceskip=0pt\relax}
\providecommand\BIBentryALTinterwordstretchfactor{4}
\providecommand\BIBentryALTinterwordspacing{\spaceskip=\fontdimen2\font plus
\BIBentryALTinterwordstretchfactor\fontdimen3\font minus
  \fontdimen4\font\relax}
\providecommand\BIBforeignlanguage[2]{{%
\expandafter\ifx\csname l@#1\endcsname\relax
\typeout{** WARNING: IEEEtran.bst: No hyphenation pattern has been}%
\typeout{** loaded for the language `#1'. Using the pattern for}%
\typeout{** the default language instead.}%
\else
\language=\csname l@#1\endcsname
\fi
#2}}

\bibitem{scherer2008}
S.~Scherer, S.~Singh, L.~J. Chamberlain, and M.~Elgersma, ``Flying fast and low
  among obstacles: Methodology and experiments,'' \emph{IJRR}, 2008.

\bibitem{dey2011cascaded}
D.~Dey, C.~Geyer, S.~Singh, and M.~Digioia, ``A cascaded method to detect
  aircraft in video imagery,'' \emph{IJRR}, 2011.

\bibitem{ross2013learning}
S.~Ross, N.~Melik-Barkhudarov, K.~S. Shankar, A.~Wendel, D.~Dey, J.~A. Bagnell,
  and M.~Hebert, ``Learning monocular reactive uav control in cluttered natural
  environments,'' in \emph{ICRA}, 2013.

\bibitem{kelly2006perceptor}
A.~Kelly \emph{et~al.}, ``Toward reliable off road autonomous vehicles
  operating in challenging environments,'' \emph{IJRR}, 2006.

\bibitem{urban2008}
M.~Buehler, K.~Iagnemma, and S.~Singh, ``Special issue on the 2007 darpa urban
  challenge, part i, ii, iii,'' \emph{JFR}, 2008.

\bibitem{knepper2009diversity}
R.~Knepper and M.~Mason, ``Path diversity is only part of the problem,'' in
  \emph{ICRA}, May 2009.

\bibitem{schmid2014autonomous}
K.~Schmid, P.~Lutz, T.~Tomi{\'c}, E.~Mair, and H.~Hirschm{\"u}ller,
  ``Autonomous vision-based micro air vehicle for indoor and outdoor
  navigation,'' \emph{JFR}, 2014.

\bibitem{matthies2014stereo}
L.~Matthies, R.~Brockers, Y.~Kuwata, and S.~Weiss, ``Stereo vision-based
  obstacle avoidance for micro air vehicles using disparity space,'' in
  \emph{ICRA}, 2014.

\bibitem{wendel2012dense}
A.~Wendel, M.~Maurer, G.~Graber, T.~Pock, and H.~Bischof, ``Dense
  reconstruction on-the-fly,'' in \emph{CVPR}, 2012.

\bibitem{srinivasan2011visual}
M.~V. Srinivasan, ``Visual control of navigation in insects and its relevance
  for robotics,'' \emph{Current opinion in neurobiology}, 2011.

\bibitem{beyeler2009vision}
A.~Beyeler, J.-C. Zufferey, and D.~Floreano, ``Vision-based control of
  near-obstacle flight,'' \emph{Autonomous robots}, 2009.

\bibitem{ardupilot}
\BIBentryALTinterwordspacing
 [Online]. Available: \url{http://dev.ardupilot.com}
\BIBentrySTDinterwordspacing

\bibitem{ros2009}
M.~Quigley, K.~Conley, B.~P. Gerkey, J.~Faust, T.~Foote, J.~Leibs, R.~Wheeler,
  and A.~Y. Ng, ``Ros: an open-source robot operating system,'' in \emph{ICRA
  Workshop on Open Source Software}, 2009.

\bibitem{coulter1992implementation}
R.~C. Coulter, ``Implementation of the pure pursuit path tracking algorithm,''
  DTIC Document, Tech. Rep., 1992.

\bibitem{saxena2005learning}
A.~Saxena, S.~H. Chung, and A.~Y. Ng, ``Learning depth from single monocular
  images,'' in \emph{NIPS}, 2005.

\bibitem{farneback2003two}
G.~Farneb{\"a}ck, ``Two-frame motion estimation based on polynomial
  expansion,'' in \emph{Image Analysis}, 2003.

\bibitem{helgason1980support}
S.~Helgason, ``Support of radon transforms,'' \emph{Advances in Mathematics},
  1980.

\bibitem{harris1998combined}
C.~Harris and M.~Stephens, ``A combined corner and edge detector.'' in
  \emph{Alvey vision conference}, 1988.

\bibitem{davies2004machine}
E.~R. Davies, \emph{Machine vision: theory, algorithms, practicalities}, 2004.

\bibitem{dalal2006human}
N.~Dalal, B.~Triggs, and C.~Schmid, ``Human detection using oriented histograms
  of flow and appearance,'' in \emph{ECCV}, 2006.

\bibitem{li2013pixel}
C.~Li and K.~M. Kitani, ``Pixel-level hand detection in ego-centric videos,''
  in \emph{CVPR}, 2013.

\bibitem{vw}
J.~Langford, L.~Li, and A.~Strehl, ``Vowpal {W}abbit,'' 2007.

\bibitem{agarwal2013least}
A.~Agarwal, S.~M. Kakade, N.~Karampatziakis, L.~Song, and G.~Valiant, ``Least
  squares revisited: Scalable approaches for multi-class prediction,''
  \emph{arXiv preprint arXiv:1310.1949}, 2013.

\bibitem{hu2014efficient}
H.~Hu, A.~Grubb, J.~A. Bagnell, and M.~Hebert, ``Efficient feature group
  sequencing for anytime linear prediction,'' \emph{arXiv:1409.5495}, 2014.

\bibitem{deyConSeqOptRSS}
D.~Dey, T.~Y. Liu, M.~Hebert, and J.~A.~D. Bagnell, ``Contextual sequence
  optimization with application to control library optimization,'' in
  \emph{RSS}, 2012.

\bibitem{batra2012diverse}
D.~Batra, P.~Yadollahpour, A.~Guzman-Rivera, and G.~Shakhnarovich, ``Diverse
  m-best solutions in markov random fields,'' in \emph{Computer Vision--ECCV
  2012}.\hskip 1em plus 0.5em minus 0.4em\relax Springer, 2012, pp. 1--16.

\bibitem{newcombe2011dtam}
R.~A. Newcombe, S.~J. Lovegrove, and A.~J. Davison, ``Dtam: Dense tracking and
  mapping in real-time,'' in \emph{ICCV}, 2011.

\bibitem{klein2007parallel}
G.~Klein and D.~Murray, ``Parallel tracking and mapping for small ar
  workspaces,'' in \emph{ISMAR}, 2007.

\bibitem{honegger2013open}
D.~Honegger, L.~Meier, P.~Tanskanen, and M.~Pollefeys, ``An open source and
  open hardware embedded metric optical flow cmos camera for indoor and outdoor
  applications,'' in \emph{ICRA}, 2013.

\bibitem{tomasi1991detection}
C.~Tomasi and T.~Kanade, \emph{Detection and tracking of point features}.\hskip
  1em plus 0.5em minus 0.4em\relax School of Computer Science, Carnegie Mellon
  Univ., 1991.

\bibitem{green2006paths}
C.~Green and A.~Kelly, ``Optimal sampling in the space of paths: Preliminary
  results,'' Robotics Institute, Pittsburgh, PA, Tech. Rep. CMU-RI-TR-06-51,
  November 2006.

\end{thebibliography}

\end{document}